\documentclass[runningheads,a4paper]{llncs}

\usepackage{amsmath,amsfonts,amssymb}
\usepackage{graphicx}
\usepackage{algorithm}
\usepackage{algorithmic}
\usepackage{multirow}

\usepackage{xcolor}
\usepackage[normalem]{ulem}

\def\bp{\mathbf{OT}}

\newcommand{\smallmat}[1]{\left[\begin{smallmatrix}#1\end{smallmatrix}\right]}

\DeclareMathOperator{\argmin}{argmin}
\DeclareMathOperator{\defeq}{\overset{\defi}{=}}
\providecommand{\abs}[1]{\left\lvert#1\right\rvert} 
\providecommand{\smabs}[1]{\lvert#1\rvert} 
\providecommand{\norm}[1]{\lVert#1\rVert}

\def\RR{\mathbb{R}}

\def\defeq{\overset{\text{def}}{=}}

\def\one{\mathbf{1}} 
\newcommand{\dotprod}[2]{\ensuremath{\langle #1 , #2\,\rangle}}

\def\slantfrac#1#2{\kern.1em^{#1}\kern-.1em/\kern-.1em_{#2}}

\begin{document}

% Guidelines :
% http://ipmi2015.cs.ucl.ac.uk/PaperSubmission/SubmissionGuidelines/index.html

% max number of pages is 12 pages

% XXX : remove for double blind submission

% \title{Geometric Averaging of Neuroimaging Data with Kantorovich Means}
\title{Fast Optimal Transport Averaging of Neuroimaging Data}

% \title{Kantorovich Means for Unormalized Measures:
%        A New Method to Average Neuroimaging Data}

% \author{X. XXX\inst{1}\inst{2}, X.
% XXX\inst{3}, X. XXX\inst{4}}

% \institute{
% XX XXXXX XXXX XX XXXX
% \and
% XX XXXXX XXXX XX XXXX
% \and
% XX XXXXX XXXX XX XXXX
% \and
% XX XXXXX XXXX XX XXXX
% }

\author{A. Gramfort\inst{1}\inst{2}, G.
Peyr\'e\inst{3}, M. Cuturi\inst{4}}

\institute{
Institut Mines-T\'el\'ecom, Telecom ParisTech, CNRS LTCI
\and
NeuroSpin, CEA Saclay, Bat. 145, 91191 Gif-sur-Yvette, cedex France
\and
CNRS and CEREMADE, Universit\'e Paris-Dauphine
\and
Graduate School of Informatics, Kyoto University
}

%%%%%%%%%%%%%%%%%%%%%%%%%%%%%%%%%%%%%%%%%%%%%%%%%%%%%%%%%%%%%%%%%%%%%%%%%%%%%%%%

\maketitle

\begin{abstract}
Knowing how the Human brain is anatomically and functionally organized
at the level of a group of healthy individuals or patients is the primary
goal of neuroimaging research. Yet computing an average of brain imaging data
defined over a voxel grid or a triangulation remains a challenge.
Data are large, the geometry of the brain is complex and the between subjects
variability leads to spatially or temporally non-overlapping effects of
interest. To address the problem of variability, data are commonly smoothed
before performing a linear group averaging. In this work we build on ideas
originally introduced by Kantorovich~\cite{kantorovich1958space} to propose a new
algorithm that can average efficiently non-normalized data defined
over arbitrary discrete domains using transportation metrics.
We show how Kantorovich means can be linked to Wasserstein barycenters
in order to take advantage of the entropic smoothing
approach used by~\cite{cuturi2014fast}. It leads to a smooth convex optimization
problem and an algorithm with strong convergence guarantees.
We illustrate the versatility of this tool and its empirical behavior
on functional neuroimaging data, functional MRI and magnetoencephalography
(MEG) source estimates, defined on voxel grids and
triangulations of the folded cortical surface.
\end{abstract}

%----------------------------------------
\section{Introduction}
%----------------------------------------

Computing the average of some observations may seem like a trivial problem,
yet it remains an active topic of research in mathematics, statistics and
applications such as medical imaging. The problem of atlas computation
from images~\cite{Joshi2004S151}, or meshes \cite{Durrleman201435},
or the problem of group analysis from functional imaging
data~\cite{thirion-etal:2007} are particularly relevant for this field.
The challenge is that natural
phenomena are usually described in terms of physical and temporal event locations,
along with their intensity. While Euclidean averaging is
standard and has some benefits such as low computation time, this
procedure ignores the geometry of the space the observations belong to;
the image of an average brain image obtained by Euclidean averaging of
individual voxels does not yield the image of the brain of an average
individual.

Starting from observations defined on a regular or irregular grid, our aim is to provide a
\emph{model-free} approach to \emph{average} them that only builds upon geometric arguments. An example of such data are functional MRI (fMRI) data defined on a voxel grid or a triangulated cortical surface. The approach aims to be intrinsically geometric in the sense that it \emph{only} requires the prior knowledge of a metric between the locations on the grid. The technique aims to be versatile in the sense that it can be applied to weighted samples
taking values on a discretized space with no assumptions on the regularity of the metric.

% XXX : \cite{Rehman2009931} PDE formulation for image registration

The approach we propose is inspired by optimal transport theory~\cite{villani09}
%,previously used for image registration~\cite{haker-etal:04}, 
and can be seen as an extension of the Wasserstein barycenter problem~\cite{agueh2011barycenters,rabin2012,cuturi2014fast}, which aims at estimating a probability measure which bests approximates a family of probability measures in the Wasserstein metric sense. The challenge of using optimal transport in the setting we consider comes from the fact that Wasserstein distances (and their barycenters) are defined for \emph{probability} measures only. While some data are normalized such the Orientation Diffusion Function (ODF) in diffusion MRI~\cite{descoteaux-etal:2009}, a number of medical imaging data are non-normalized. Here we bypass this limitation using a generalization of optimal transport distances proposed by~\cite{kantorovich1958space}. This extension comes at a price, since it introduces an additional parameter (the cost of adding or removing mass) which can be difficult to tune. We propose a simple way to mitigate this problem by introducing a natural constraint on the overall mass of the barycenter, which we set to be equal to the average of the masses of all samples. We provide an efficient method to compute Kantorovich means by building upon the first algorithm of~\cite{cuturi2014fast}. We provide intuitions on the behavior of our method and demonstrate its relevance with simulations and experimental results obtained with fMRI and MEG data which are neuroimaging data defined on a voxel grid or a triangulation of the folded cortical mantle.

This paper is organized as follows. We start in Section~\ref{sec:background} with a reminder on optimal transport and the Kantorovich metric for non-normalized measures. We introduce Kantorovich means in Section~\ref{sec:contrib} and describe efficient algorithms to compute them. Section~\ref{sec:experiments} contains simulations and results on publicly available fMRI data with 20 subjects and MEG data with 16 subjects.

\def\pO{\RR^{d}_+}
%----------------------------------------
\section{Optimal Transport Between (Un)Normalized Measures}\label{sec:background}
%----------------------------------------
We introduce in this section Wasserstein distances for non-negative measures on a finite metric space $(\Omega,D)$. Simply put, we consider normalized histograms on a grid of locations, and assume the distance between any two locations is provided. Next, we extend Wasserstein distances to non-negative vectors of \emph{bounded mass}.

\paragraph{Notations.} Let $d$ be the cardinal of $\Omega$. Relabeling arbitrarily all the elements of $\Omega$ as $\{1,\cdots, d\}$ we represent the set of non-negative measures on $\Omega$ by the non-negative orthant $\pO$. Let $\one_d$ be the $d$-dimensional vector of ones. For any vector $u\in\RR^d$ we write $\smabs{u}_1$ for the $l_1$ norm of $u$, $\sum_{i=1}^d \smabs{u_i}$. When $u\in\pO$ we also call $\smabs{u}_1$ the (total) \emph{mass} of vector $u$. Let $M=[m_{ij}]\in\RR^{d\times d}_+$ be the matrix that describes the metric between all locations in $\Omega$, namely $m_{ij}=D(i,j), i,j\leq d$. %Note that, because of the metric axioms, we necessarily have for all $i,j,k\leq d$ that $m_{ij}\leq m_{ik}+m_{kj}$.  % we don't need this.

\paragraph{Wasserstein Distance for Normalized Histograms.}
Consider two vectors $a,b\in\pO$ such that $\smabs{a}_1=\smabs{b}_1$. Both can be interpreted as histograms on $\Omega$ of the same mass. A non-trivial example of such normalized data in medical imaging is the discretized ODF used for diffusion imaging data~\cite{descoteaux-etal:2009}. For $p\geq 1$, the $p$-Wasserstein distance $W_p(a,b)$ between $a$ and $b$ is the $p^\text{th}$ root of the optimum of a linear program, known as a \emph{transportation problem} \cite[\S7.2]{bertsimas1997introduction}. A transport problem is a network flow problem on a bipartite graph with cost $M^p$ (the pairwise distance matrix $M$ raised element-wise to the power $p$), and feasible set of flows $U(a,b)$ (known as the transportation polytope of $a$ and $b$), where $U(a,b)$ is the set of $d\times d$ nonnegative matrices such that their row and column marginals are equal to $a$ and $b$ respectively:
\begin{equation}\label{eq:polytope} 
	U(a,b) \defeq \{ T\in\mathbb{R}_+^{d\times d}\; |\; T\one_d = a,\, T^T\one_d = b \}.
\end{equation}
Given the constraints induced by $a$ and $b$, one naturally has that $U(a,b)$ is empty when $\smabs{a}_1\ne\smabs{b}_1$ and non-empty when $\smabs{a}_1=\smabs{b}_1$ (in which case one can easily check that the matrix $ab^T/\smabs{a}_1$ belongs to that set). The $p$-Wasserstein distance $W_p(a,b)$ raised to the power $p$ (written $W^p_p(a,b)$ below) is equal to the optimum of a parametric Optimal Transport ($\bp$) problem on $d^2$ variables,
\begin{equation}\label{eq:primal}W_p^p(a,b) = \bp(a,b,M^p)\defeq\min_{T\in U(a,b)}\dotprod{T}{M^p},\end{equation}
parameterized by the marginals $a,b$ and matrix $M^p$. 
%Note that if either $a$ or $b$ do not have full support, the size of that program can be trivially reduced to be equal to the product of the sizes of the support of $a$ and $b$. We leave aside such details in the rest of this paper, since they only play a role in the algorithms' implementation, but not on their mathematical properties.

\paragraph{Optimal Transport for Unnormalized Measures.} If the total masses of $a$ and $b$ differ, namely $\smabs{a}_1\ne\smabs{b}_1$, the definition provided above is not useful because $U(a,b)=\emptyset$. Several extensions of the OT problem have been proposed in that setting; we recall them here for the sake of completeness. In the computer vision literature, \cite{rubner1997earth} proposed to handle that case by: 
\emph{(i)} \emph{relaxing}
 the equality constraints of $U(a,b)$ to inequality constraints $T\one_d \leq a,\, T^T\one_d \leq b$ in Equation~\eqref{eq:polytope}; \emph{(ii)} \emph{adding an equality constraint} on the total mass of the solution $\one_d^TT\one_d=\min(\smabs{a}_1,\smabs{b}_1)$; \emph{(iii)} \emph{dividing} the minimum of $\dotprod{T}{M}$ under constraints \emph{(i,ii)} by $\min(\smabs{a}_1,\smabs{b}_1)$. This modification does not, however, result in a metric. \cite{pele2008linear} proposed later a variant of this approach called EMD-hat that incorporates constraints \emph{(i,ii)} but \emph{(iii')} adds to the optimal cost $\dotprod{T^\star}{M}$ a constant times $\min(\smabs{a}_1,\smabs{b}_1)$. When that constant is large enough $M$, \cite{pele2008linear} claim that EMD-hat is a metric. We also note that \cite{benamou2003numerical} proposed a quadratic penalty between the differences of masses and made use of a dynamic formulation of the transportation problem. % in the field of computational fluid mechanics.

\paragraph{Kantorovich Norms for Signed Measures.} We propose to build on early contributions by Kantorovich to define a generalization of optimal transport distance for unnormalized measures, making optimal transport applicable to a wider class of problems, such as the averaging of functional imaging data. \cite{kantorovich1958space} proposed such a generalization as an intermediary result of a more general definition, the \emph{Kantorovich norm} for signed measures on a compact metric space, which was itself extended to separable metric spaces by~\cite{hanin1999extension}. We summarize this idea here by simplifying it to the case of interest in this paper where $\Omega$ is a finite (of size $d$) probability space, in which case signed measures are equivalent to vectors in $\RR^d$. \cite{kantorovich1958space} propose first a norm for vectors $z$ in the orthogonal of $\one_d$ (vectors $z$ such that $z^T\one_d=0$), by considering the 1-Wasserstein distance between the positive and negative parts of $z$, $\norm{z}_K= W_1(z_+,z_-)$. A penalty vector $\Delta\in\pO$ is then introduced to define the norm $\norm{x}_{K}$ of \emph{any} vector $x$ as the minimal value of $\norm{z}_K + \Delta^T\smabs{z-x}$ when $z$ is taken in the space of all vectors $z$ with zero sum, and $\smabs{z-x}$ is the element-wise absolute value of the difference of vectors $z$ and $x$. 
For this to define a true norm in $\RR^d$, $\Delta$ must be such that $\Delta_i\geq \max_j m_{ij}$ and $\smabs{\Delta_i-\Delta_j}\leq m_{ij}$. The distance between two arbitrary non-negative vectors $a,b$ of different mass is then defined as $\norm{a-b}_K$. 
As highlighted by~\cite[p.108]{villani09}, and if we write $e_i$ for the $i^{\text{th}}$ vector of the canonical basis of $\RR^d$, this norm is the maximal norm in $\RR^d$ such that for any $i,j \leq d$, $\norm{e_i-e_j}_K=m_{ij}$, namely the maximal norm in the space of signed measures on $\Omega$ such that the norm between two Dirac measures coincides with $\Omega$'s metric between these points.

\paragraph{Kantorovich Distances for Unnormalized Nonnegative Measures.} \cite{guittet2002extended} noticed that Kantorovich's distance between unnormalized measures can be cast as a regular optimal transport problem. Indeed, one simply needs to: \emph{(i)} add a \emph{virtual point} $\omega$ to the set $\Omega=\{1,\cdots,d\}$ whose distance $D(i,\omega)=D(\omega,i)$ to any element $i$ in $\Omega$ is set to $\Delta_i\,$; \emph{(ii)} use that point $\omega$ as a buffer when comparing two measures of different mass. The appeal of Kantorovich's formulation in the context of this work is that it boils down to a classic optimal transport problem, which can be approximated efficiently using the smoothing approach of \cite{cuturi2013sinkhorn} as discussed in Section~\ref{sec:contrib}. To simplify our analysis in the next section, we only consider non-negative vectors (histograms) $a\in\pO$ such that their total mass is upper bounded by a known positive constant. This assumption alleviates the definition of our distance below, since it does not require to treat separately the cases where either $\smabs{a}_1 \geq \smabs{b}_1$ or $\smabs{a}_1 < \smabs{b}_1$ when comparing $a,b\in\pO$. Note also that this assumption always holds when dealing with finite collections of data. Without loss of generality, this is equivalent to considering vectors $a$ in $\pO$ such that $\smabs{a}_1\leq 1$ with a simple rescaling of all vectors by that constant. We define next the Kantorovich metric on $S_d$, where $S_d=\{u\in\pO, \smabs{u}_1 \leq 1\}$.

\begin{definition}[Kantorovich Distances on $S_d$] Let $\Delta\in\pO$ such that $\Delta_i\geq \max_j m_{ij}$ and $\smabs{\Delta_i-\Delta_j}\leq m_{ij}$. Let $p\geq 0$. For two elements $a$ and $b$ of $S_d$, we write $\alpha= 1-\smabs{a}_1\geq 0$ and  $\beta= 1-\smabs{b}_1\geq 0$. Their $p$-Kantorovich distance raised to the power $p$ is
\begin{equation}\label{eq:kantodis}K^p_{p \Delta}(a,b)= \bp(\smallmat{a\\\alpha},\smallmat{b\\\beta},\hat{M}^p), \text{ where } \hat{M}=\begin{bmatrix}M & \Delta\\\Delta^T & 0\end{bmatrix}\in\RR_+^{d+1\times d+1}.\end{equation}
\end{definition}
The Kantorovich distance inherits all metric properties of Wasserstein distances: the mapping which to a vector $a$ associates a vector $[a;1-\smabs{a}_1]\in\Sigma_{d+1}$ can be regarded as a feature map, to which the standard Wasserstein distance using $\hat{M}$ (which is itself a metric matrix) is applied.

%\todo[inline]{mention that this is equivalent to a dual weight on $1-\sum_i a_i$}

%----------------------------------------
\section{Kantorovich Mean of Unnormalized Measures}\label{sec:contrib}
%----------------------------------------
Consider now a collection $\{b^1,\cdots, b^N\}$ of $N$ non-negative measures on $(\Omega,D)$ with mass upper-bounded by $1$, namely $N$ vectors in $S_d$. Let $\beta^j=1-\abs{b^j}$ be the deficient mass of $b^j$. Our goal in this section is to find, given a vector of virtual costs $\Delta$ and an exponent $p$, a vector $a$ in $S_d$ which minimizes the sum of its $p$-Kantorovich distances $K^p_{p \Delta}$ to all the $b^j$,
\begin{equation}
    \label{eq:originalkantomean}\tag{P1}
    a \in \underset{u\in S_d}{\argmin} \frac{1}{N}\sum_{j=1}^N K_{p \Delta}^p(u,b^j)
    =\underset{u\in S_d}{\argmin} \frac{1}{N}\sum_{j=1}^N \bp(\smallmat{u\\1-\smabs{u}_1},\smallmat{b^j\\\beta^j},\hat{M}^p).
\end{equation}
Because of the equivalence between Kantorovich distances for points in $S_d$ and Wasserstein distances in the $d+1$ simplex, this problem can be naturally cast as a Wasserstein barycenter problem~\cite{agueh2011barycenters} with metric $\hat{M}$. Problem (P1) can be cast as a linear program with $N(d+1)^2$ variables. For the applications we have in mind, where $d$ is of the order or larger than $10^4$, solving that program is not tractable. We discuss next computational approaches to solve it efficiently.

\paragraph{Smooth Optimal Transport.} \cite{rabin2012} and \cite{BRPP13} have proposed efficient algorithms to solve the Wasserstein barycenter problem in low dimensional Euclidean spaces. These approaches are not, however, suitable when one considers observations on the cortex, for which \emph{all pairs shortest path} metrics (inferred from a graph structure connecting all voxels) are preferred over Euclidean metrics. To solve Problem~\eqref{eq:originalkantomean} we turn instead to a recent series of algorithms proposed in \cite{cuturi2014fast}, \cite{benamou2014iterative} and \cite{cuturi2015smoothed} that all exploit the regularized OT approach suggested in \cite{cuturi2013sinkhorn}. Among these recent approaches, we propose to build in this work upon the first algorithm in \cite{cuturi2014fast}, which can be easily modified to incorporate constraints on $a$. This flexibility will prove useful in the next section. 

The strategy of \cite{cuturi2014fast} is to regularize directly the optimal transport problem by an entropic penalty, whose weight is parameterized by a parameter $\lambda>0$,
\[
    \bp_\lambda(a,b,M^p)\defeq\min_{T\in U(a,b)}\dotprod{T}{M^p}-\frac{1}{\lambda} H(T),
\]
where $H(T)$ stands for the entropy of the matrix $T$ seen as an element of the simplex of size $d^2$,
$H(T) \defeq - \sum_{ij} t_{ij} \log (t_{ij})$.
As shown by \cite{cuturi2014fast}, the regularized transport problem $\bp_\lambda$ admits a unique optimal solution. As such, $\bp_\lambda(a,b,M^p)$ is a differentiable function of $a$ whose gradient can be recovered through the solution of the corresponding smoothed dual optimal transport. Without elaborating further on this approach, we propose to simply replace all expressions that involve an optimal transport problem $\bp$ in our formulations by their smoothed counterpart $\bp_\lambda$.% and adapt Algorithm 1 in \cite{cuturi2014fast}.

\paragraph{Sensitivity of Kantorovich Means to the Parameter $\Delta$.} The magnitude of the solution $a$ to Problem~\eqref{eq:originalkantomean} depends directly on the virtual distance $\Delta$. Suppose, for instance, that $\Delta=\varepsilon \one_d$ with $\varepsilon$ arbitrarily small. In that case $a$ should converge to a unit mass on the last (virtual) bin and would therefore be equal to the null histogram $\mathbf{0}_d$ on the $d$ other bins. If, on the contrary, $\Delta=\gamma\one_d$ and $\gamma$ is large, we obtain that $K_{p\Delta}^p(a,b)/\gamma$ grows as $\abs{\,\smabs{a}_1-\smabs{b}_1}$. Therefore a minimum of Problem~\eqref{eq:originalkantomean} would necessarily need to have a total mass that minimizes $\sum_j\abs{\,\smabs{a}_1-\smabs{b^j}_1}$, namely a total mass equal to the median mass of all $b^j$.
% The sensitivity of the barycenter to the values of $\Delta$ is illustrated in the upper-middle plot of Figure~\ref{fig:pcurves}.
This sensitivity of the solution $a$ to the magnitude of $\Delta$ may be difficult to control. Choosing adequate values for $\Delta$, namely setting the distance of the virtual point to the $d$ other points, may also be a difficult parameter choice. To address this issue we propose to simplify our framework by introducing an equality constraint on the mass of the barycenter $a$ in our definition, and let $\Delta$ be any non-negative vector, typically set to a large quantile of the distribution of all pairwise distances $M_{ij}^p$ times the vector of ones $\one_d$. Under these assumptions, we can now propose $p$-Kantorovich means with a constraint on the total mass of $a$. Remaining parameters in our approach are therefore only $p$ and $\lambda$. In practice we will fix $p=1$, which corresponds to the Earth Mover's Distance~\cite{rubner1997earth}, and use a high $\lambda$, namely a small entropic regularization of order $1/\lambda$, which has also the merit of making Problem~\eqref{eq:originalkantomean} strongly convex. $\lambda$ is set in our experiments to $100/\text{median}(M)$, where $\text{median}(M)$ is the median of all pairwise distances $\{M_{ij}\}_{ij}$.

\begin{definition}[$p$-Kantorovich Means with Constrained Mass] Let $\Delta\in\pO$ and $p\geq 0$. A Kantorovich mean with a target mass $\rho\leq 1$ of a set of $N$ histograms $\{b^1,\cdots, b^N\}$ in $S_d$ is the unique vector $a$ in $S_d$ such that:
$$a\in \underset{\substack{a\in S_d\\\smabs{a}_1=\rho}}{\argmin}\frac{1}{N}\sum_j \bp_\lambda(a,b^j,\hat{M}^p).$$ 
\end{definition}
We provide in Algorithm~\ref{algo:kantor} an implementation of~\cite[Alg.1]{cuturi2014fast}. Unlike their version, we only consider a fixed step-length exponentiated gradient descent, and add a mass renormalization step. We set the default mass of the barycenter to be the mean of the masses of all histograms. We use the notation $\circ$ for the elementwise (Schur) product of vectors. Note that the computations of $N$ dual optima in line 7 of Algorithm~\ref{algo:kantor} below can be vectorized and computed using only matrix-matrix products. We use GPGPUs to carry out these computations.

\begin{algorithm}
	\begin{algorithmic}[1]
		\caption{$p$-Kantorovich Barycenter with Constrained Mass\label{algo:kantor}} 
		\STATE \textbf{\emph{Input}}: $\{b^1,\cdots,b^N\} \subset S_d$, metric $M$, quantile $q$, $p\geq 0$, entropic regularizer $\lambda>0$, step size $c$.
		\STATE Compute mean mass $\rho=\frac{1}{N}\sum_i \smabs{b^j}_1$.
		\STATE Form virtual cost vector $\Delta=\text{quantile}(M,q\%)\one_d$.
		\STATE Form augmented $d+1\times d+1$ ground metric $\hat{M}$ as in Equation~\eqref{eq:kantodis}
		\STATE Set $a=\one_{d+1}/(d+1).$
		\WHILE{$a$ changes}
		\STATE Compute all dual optima $\alpha^j$ of $\bp_\lambda(a,b^j,\hat{M})$ using \cite[Alg.3]{cuturi2014fast}
		\STATE $a\leftarrow a \circ \exp(-c \frac{1}{N}\sum_{j=1}^N \alpha^j)$; (gradient update)
		\STATE $a_i \leftarrow \begin{cases}\rho a_i/\sum_{l=1}^d a_l &\text{ if } i\leq d, \\
1-\rho &\text{ if } i=d+1.\end{cases}$ (projection on the simplex/mass constraint)
		\ENDWHILE
		\STATE \textbf{\emph{Output}} $a_{1:d}\in S_d$.
	\end{algorithmic}
\end{algorithm}

% \vspace{-.4cm}
%%%%%%%%%%%%%%%%%%%%%%%%%%%%%%%%
\section{Application to the Averaging of Neuroimaging Data}\label{sec:experiments}
%%%%%%%%%%%%%%%%%%%%%%%%%%%%%%%%

% The study of neural mechanisms and brain architecture using imaging technologies
% such as magnetic resonance imaging (MRI), functional MRI (fMRI),
% magnetoencephalography (MEG) and
% electroencephalography (EEG) is called neuroimaging.
% Knowing how the Human brain is anatomically and functionally organized
% at the level of a group of healthy individuals or patients
% is the primary goal of brain research. The problem of
% averaging data at the group level is therefore central.

% We now present results on functional brain imaging data.

Neuromaging data are defined on a grid of voxels, eventually
restricted to the brain volume, or on a triangulation of the cortical mantle
obtained by segmentation of MRI data.
Examples of data most commonly analyzed on a grid are fMRI data,
while neural activity estimates derived from MEG/EEG data are often
restricted to the cortical surface~\cite{dspm}. Anatomical data such as
cortical thickness, which is a biomarker of certain neurodegenerative pathologies,
is also defined on the surface. In all cases the data are defined
on a discrete set of points (voxels or vertices) which have a
natural distance given by the geometry of the brain. The data are
also non-normalized as they represent physical or statistical
quantities, such as thickness in millimeters or F statistics.
Such data are therefore particularly well adapted to the algorithm
proposed in this paper: they are defined on discrete space, are non-normalized
and there exists a natural ground metric.

The difficulty when averaging neuroimaging
data is the anatomo-functional variability: every brain is different.
The standard approach to compensate for this variability
across subjects is to smooth the data to favor the
overlap of signal of interest after the individual data have been
ported to a common space using anatomical landmarks
(anatomical registration). Volume data are typically redefined
in MNI space while surface data are transferred to an average
cortical surface, using for instance the FreeSurfer
software\footnote{\url{https://surfer.nmr.mgh.harvard.edu/}}. When working
with fMRI the spatial variability is commonly compensated
by smoothing the data with an isotropic Gaussian kernel of Full Width at Half Maximum (FWHM) between 6 and
8\,mm. MEG and EEG suffer from the same spatial variability, but also
from the temporal variability of neural responses which is compensated
by low pass filtering the data. By employing a transportation metric
informed by the geometry of the domain, this smoothing procedure
as well as the setting of the kernel bandwidth are not needed.

In the following experiments we focus on spatial
averaging, although extension to spatiotemporal data is straightforward
provided the metric is defined along the time axis.
When working with a voxel grid the distance is the Euclidian
distance taking into account the voxel size in millimeters, and
when working on a cortical triangulation, the distance used
is the geodesic distance computed on the folded cortical mantle.
We now present results of a simulation study where standard averaging
with Gaussian smoothing is compared to Kantorovich means.
Simulation results are followed by experimental results obtained
with fMRI data from 20 subjects and MEG data
on a population of 16 subjects.

% \cite{scherg-etal:85}

\paragraph{Simulation setup.}

% In a first simulation, we mimick fMRI data analyzed on a voxel grid.
% This simulation presented in Fig.~\ref{fig:simu_grid_results} aims to illustrate
% the behavior of our algorithm in a classical context of functional
% brain imaging. In Fig.\ref{fig:simu_grid_results}-c we report
% results obtained by classical averaging while Fig.\ref{fig:simu_grid_results}-d
% shows the Kantorovich barycenter with constrained mass. One can observe that
% this barycenter highlights a clear active
% region without requiring any kernel smoothing.

% \begin{figure}[t!]
%     \centering
%         \includegraphics[width=0.24\linewidth]{simu_voxel_data.png}
%         \includegraphics[width=0.24\linewidth]{simu_voxel_data_crop.png}
%         \includegraphics[width=0.24\linewidth]{simu_voxel_data_mean.png}
%         \includegraphics[width=0.24\linewidth]{simu_voxel_data_ot.png}
%     \caption{Illustration on a voxel grid with simulated fMRI data.
%     The data consist
%     of three subjects with for each a small cluster of active voxels.
%     There is no spatial overlap between the three clusters to make them
%     distinguishable.
%     From left to right, a) the sum of the three volumes without smoothing,
%     b) the same as a) with zoom on region of interest, c) the standard average
%     following Gaussian smoothing, d) the Kantorovich mean with Euclidian
%     ground metric and p=1.
%     The later result highlights a clear foci of activations
%     in the ROI without smearing the activation nor
%     damping the amplitudes as much as the kernel smoothing.}
%     \label{fig:simu_grid_results}
% \end{figure}

In this experiment using on a triangulation of the cortex,
we simulated signals of interest in two brain regions using the functional
parcellation provided by the FreeSurfer software. We used regions
Broadman area 45 (BA45) and the visual area MT. We simulated for a group
of 100 subjects random positive signals in these two regions. For each
subject and each region, the signal is focal at a random location
with a random amplitude generated with a truncated Gaussian distribution
(mean 5, std. dev. 1.). We use here focal signals to exemplify
the effect of optimal transport. Such signals could correspond to dipolar
activations derived from MEG/EEG using dipole fitting methods~\cite{scherg-etal:85}
or sparse regression techniques~\cite{wipf-etal:07,gramfort-etal:2013}.

\begin{figure}[t!]
    \centering
        \includegraphics[width=0.6\linewidth]{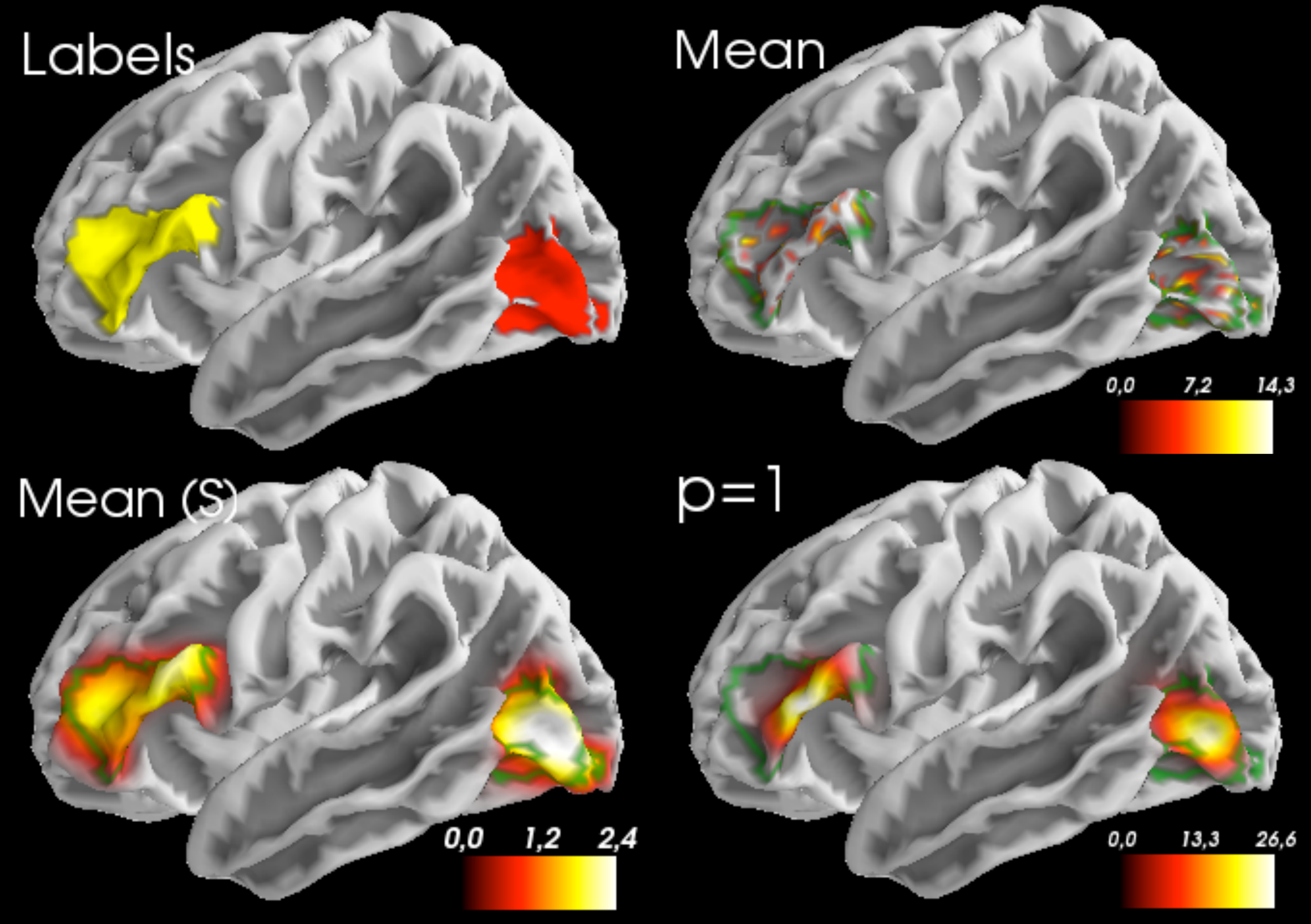}
    \caption{Simulation results with focal random signals generated in areas/labels BA45 (yellow) and MT (red)
    in a group of 100 subjects. Data are defined on a surface with 10,024 vertices
    (FreeSurfer fsaverage 5).
    One shows the standard averaging referred to as \emph{Mean},
    the averaging after Gaussian smoothing is referred to as \emph{Mean (S)}
    (mean after Gaussian smoothing with FWHM=8\,mm),
    and the Kantorovich mean (p=1).
    The result \emph{Mean} shows the focal signals with
    random positions in the labels delineated in green. The Kantorovich mean
    highlights clear foci of activations in the ROIs without
    smearing the activation as with Gaussian smoothing which furthermore
    significantly dampens the amplitudes.}
    \label{fig:simu_results}
\end{figure}

Figure~\ref{fig:simu_results} presents the locations of the two
regions (labels), the averages with and without Gaussian
smoothing and the Kantorovich average. 
Gaussian smoothing leads to a highly
blurred average which exceeds the extent of the regions of interest,
while it also strongly reduces the amplitudes of the signals,
potentially washing out the statistical effects.
The peak amplitudes obtained with optimal transport are also
higher and closer to the individual peak amplitudes.
One can clearly observe the limitations of Gaussian smoothing,
which furthermore requires to set the bandwidth of the kernel.
The Kantorovich average nicely highlights two foci of signals
at the group level.

\paragraph{Results on fMRI data.}

\begin{figure}[t!]
    \centering
        \includegraphics[width=0.27\linewidth]{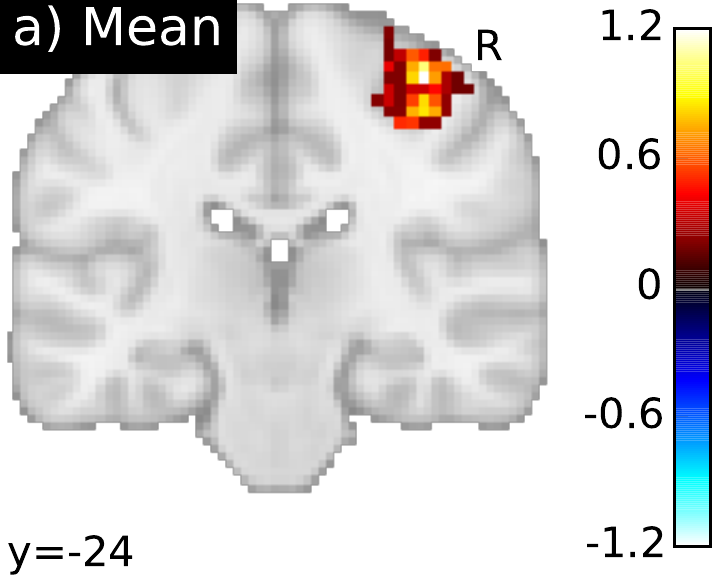}
        \includegraphics[width=0.27\linewidth]{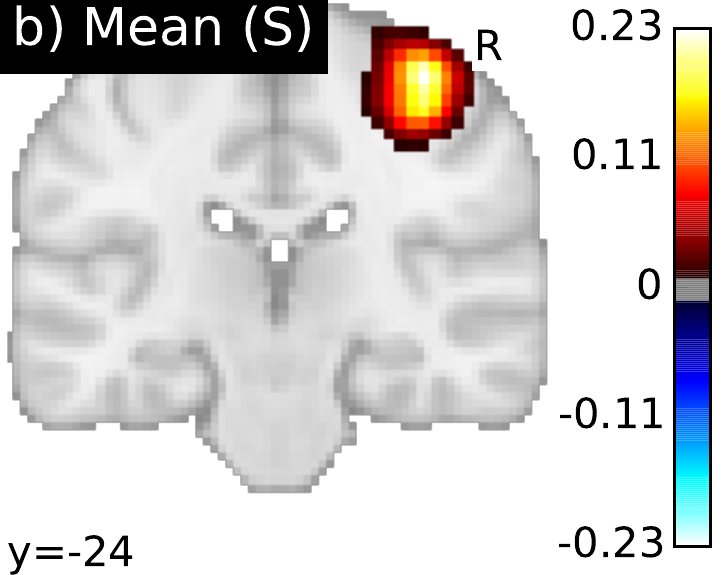}
        \includegraphics[width=0.27\linewidth]{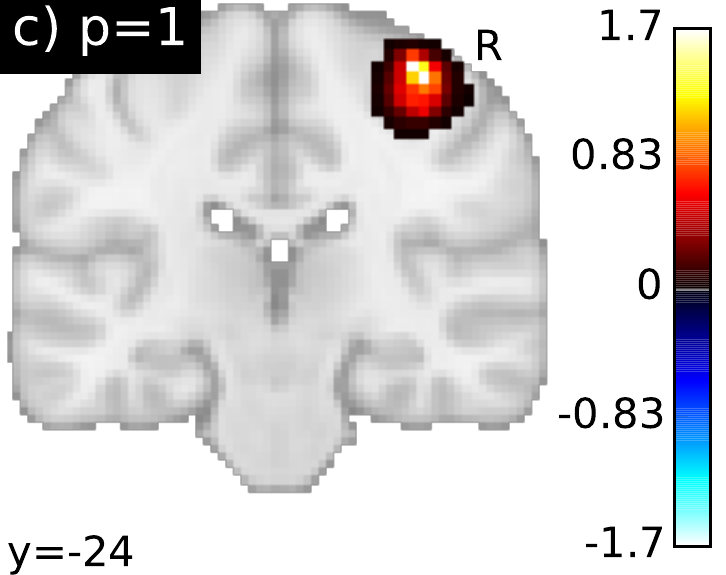}
    \caption{Averaging of the standardized effect of interest on fMRI
    data.
    From left to right, a) the Euclidian mean without smoothing,
    b) the Euclidian mean with smoothing (FWHM=8\,mm), c)
    the Kantorovich mean with Euclidian ground metric and p=1.
    The later result highlights a clear foci of activations
    in the ROI without smearing the activation nor
    damping the amplitudes as much as the kernel smoothing.}
    \label{fig:fmri_grid_results}
\end{figure}

We used here fMRI data analyzed on a voxel grid.
It corresponds to 20 subjects from the database described in~\cite{pinel2007}.
We average here the standardized effect of interest induced by left hand button press.
In Fig.~\ref{fig:fmri_grid_results}-a we show the Euclidian average without
smoothing.
In Fig.\ref{fig:fmri_grid_results}-b we report
results obtained by classical averaging following Gaussian smoothing
with FWHM of 8\,mm. Fig.\ref{fig:fmri_grid_results}-c
shows the Kantorovich mean with constrained mass. One can observe that
this barycenter highlights a clear active
region without requiring any kernel smoothing. It also leads
to a amplitude in the average standardized effect around 1.7 which
is much higher than the 0.23 obtained when smoothing.

\paragraph{Results on MEG data.} We now evaluate the benefit of the proposed
approach on experimental data. These data were acquired
with a Neuromag VectorView system (Elekta Oy, Helsinki, Finland)
with 306 sensors arranged in 102 triplets, each comprising two orthogonal
planar gradiometers and one magnetometer. Subjects are presented with images
containing faces of familiar (famous)
or unfamiliar persons and so called ``scrambled'' faces.
See~\cite{henson-etal:11} for more details.
Dataset contains 16 subjects. For each one, event related fields (ERF) were
obtained by averaging about 200 repetitions of recordings following
stimuli presentations. Data were band pass filtered between 1 and 40\,Hz.
Following standard MEG source localization pipelines~\cite{mne}, a noise covariance
was estimated from prestimulus time intervals and used for
source reconstruction with the cortically constrained
dSPM method~\cite{dspm}.
The values obtained with dSPM can be considered as F statistics,
where high values are located in active regions.

\begin{figure}[h!]
    \centering
        \includegraphics[width=0.33\linewidth]{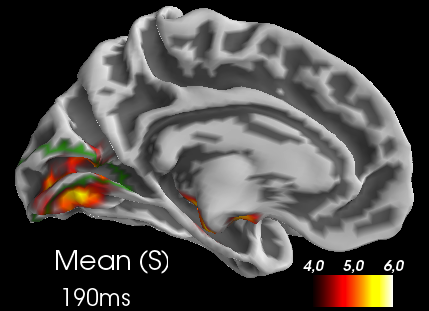}
        \includegraphics[width=0.33\linewidth]{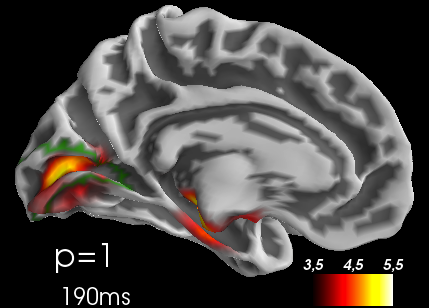}

    \caption{Average of dSPM estimates derived from MEG ERF data on a group of 16 subjects stimulated with
    pictures of famous faces. From left to right: standard mean and Kantorovich mean.
    The left hemisphere is displayed in medial view.
    In green is the border of the primary visual cortex (V1) provided by FreeSurfer.
    One can observe that the Kantorovich mean has its peak amplitude within V1.}
    \label{fig:meg_results}
\end{figure}

\begin{figure}[h!]
    \centering
        \includegraphics[width=0.325\linewidth]{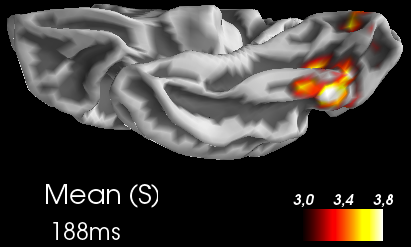}
\includegraphics[width=0.325\linewidth]{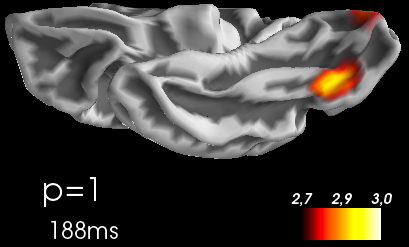}

    \caption{Group averages (16 subjects) of dSPM estimates derived from MEG
    ERF data obtained by contrasting the famous faces stimulation with the scrambled faces.
    From left to right: standard mean and the Kantorovich mean.
    The right hemisphere is displayed in ventral view. Optimal transport results highlight a focal
    activity in the Fusiform gyrus known to be implicated in face
    processing~\cite{kanwisher-etal:97}.}
    \label{fig:meg_results_ffa}
\end{figure}

In Figure~\ref{fig:meg_results}, we present results at a single
time point, 190\,ms after stimulus onset,
which corresponds to the time instant where the dSPM amplitudes
are maximum. Data correspond the visual presentation of
\emph{famous faces}.
In green, is the border of the primary visual
cortex (V1) provided by the FreeSurfer functional atlas.
One can observe that the Kantorovich barycenter yields
a more focal average nicely positioned in the middle of
the calcarine fissure where V1 is located.
Such a strong activation in V1 is expected in
such an experiment consisting of visual stimuli.
To investigate more subtle cognitive effects, such as the
response of the fusiform face area (FFA) reported
about 170\,ms after stimulation in the
literature~\cite{kanwisher-etal:97,henson-etal:11}, we
report results obtained on contrasts of ERFs measured after famous faces
presentations \emph{vs.} scrambled faces. As illustrated in
Figure~\ref{fig:meg_results_ffa}, Kantorovich mean
nicely delineates a focal source of activity in the ventral
part of the cortex known as the fusiform gyrus.

These results show that Kantorovich means provides focal
activation at the population level despite the challenging
problem of inter-subject anatomo-functional variability.
They avoid the smearing of the signal or statistical effects of interests
which naturally occur when data are spatially
smoothed before standard averaging.
Note again that here no smoothing parameter with FWHM in millimeters
is manually specified. Their solution only depends on the metric
naturally derived from the geometry
of the cortical surface.
With a cortical triangulation containing 10,024 vertices and 16 subjects
the computation on a Tesla K40 GPU of one barycenter takes less than
1\,min.

% \vspace{-0.4cm}

\section{Conclusion}

The contributions of this work are two-fold.
First, by considering non-normalized measures particularly
relevant for medical imaging data
we extend the current state of the art in barycenter estimation
using transportation metrics.
Following recent contributions on discrete optimal transport
we propose a smoothed version of the transport problem
that leads us to an efficient optimization algorithm.
While many contributions on optimal transport work only in one or two
dimensions on a regular grid, our approach can cope with the
complex geometry of the brain (irregular grids and surfaces).
Only the definition of a ground metric is here required.
The algorithm proposed involves simple operations
that are particularly adapted to modern GPU hardware and allows
us to compute barycenters on full brain data in a few minutes.

Second, with simulations defined on the cortex triangulation,
a publicly available fMRI dataset with 20 subjects
and an MEG dataset processed with a standard analysis pipeline
with 16 subjects we demonstrated the
ability of the method to clearly highlight activation foci
while avoiding the need to smooth the data. The fMRI data
showed a clear activation in the right motor cortex and
on the MEG data we showed that the proposed approach better
identified activation foci in the primary visual cortex and the
fusiform gyrus. Both findings, that are consistent with previous
neuroscience literature, show that method proposed yields
more accurate results than the current pipelines which
furthermore requires to set a kernel bandwidth
parameter. The removal of any free parameter in the pipeline
is a way towards more reproducible neuroimaging results.

Due to the non-linearity of the approach the estimation of
statistical threshold shall be performed with non-parametric
permutation tests. When thresholding barycenters as presented
in Section~\ref{sec:experiments} it is expected that one will
obtain clear clusters.

\paragraph{Acknowledgements} A. Gramfort was supported by the ANR grant
THALAMEEG, ANR-14-NEUC-0002-01. M. Cuturi gratefully acknowledges the support of JSPS young researcher A grant 26700002, the gift of a K40 card from NVIDIA and fruitful discussions with K.R. M\"uller. The work of G. Peyr\'e has been supported by the European Research Council (ERC project SIGMA-Vision).

\bibliographystyle{splncs03}
\bibliography{bib_short,bib_neuro}

\begin{thebibliography}{10}
\providecommand{\url}[1]{\texttt{#1}}
\providecommand{\urlprefix}{URL }

\bibitem{agueh2011barycenters}
Agueh, M., Carlier, G.: Barycenters in the {W}asserstein space. SIAM Journal on
  Mathematical Analysis  43(2),  904--924 (2011)

\bibitem{benamou2003numerical}
Benamou, J.D.: Numerical resolution of an “unbalanced” mass transport
  problem. ESAIM: Mathematical Modelling and Numerical Analysis  37(5),
  851--868 (2003)

\bibitem{benamou2014iterative}
Benamou, J.D., Carlier, G., Cuturi, M., Nenna, L., Peyr{\'e}, G.: Iterative
  bregman projections for regularized transportation problems. arXiv preprint
  arXiv:1412.5154  (2014)

\bibitem{bertsimas1997introduction}
Bertsimas, D., Tsitsiklis, J.: {Introduction to linear optimization}. Athena
  Scientific Belmont, MA (1997)

\bibitem{BRPP13}
Bonneel, N., Rabin, J., Peyr\'e, G., Pfister, H.: Sliced and radon wasserstein
  barycenters of measures. Journal of Mathematical Imaging and Vision pp. 1--24
  (2014)

\bibitem{cuturi2013sinkhorn}
Cuturi, M.: Sinkhorn distances: Lightspeed computation of optimal transport.
  In: Advances in Neural Information Processing Systems 26. pp. 2292--2300
  (2013)

\bibitem{cuturi2014fast}
Cuturi, M., Doucet, A.: Fast computation of wasserstein barycenters. In:
  Proceedings of the 31st International Conference on Machine Learning
  (ICML-14) (2014)

\bibitem{cuturi2015smoothed}
Cuturi, M., Peyr{\'e}, G., Rolet, A.: A smoothed dual approach for variational
  wasserstein problems. arXiv preprint arXiv:1503.02533  (2015)

\bibitem{dspm}
Dale, A., Liu, A., Fischl, B., Buckner, R.: Dynamic statistical parametric
  neurotechnique mapping: combining f{MRI} and {MEG} for high-resolution
  imaging of cortical activity. Neuron  26,  55--67 (2000)

\bibitem{descoteaux-etal:2009}
Descoteaux, M., Deriche, R., Knosche, T., Anwander, A.: Deterministic and
  probabilistic tractography based on complex fibre orientation distributions.
  Medical Imaging, IEEE Transactions on  28(2),  269--286 (Feb 2009)

\bibitem{Durrleman201435}
Durrleman, S., Prastawa, M., Charon, N., Korenberg, J.R., Joshi, S., Gerig, G.,
  Trouv\'e, A.: Morphometry of anatomical shape complexes with dense
  deformations and sparse parameters. NeuroImage  101(0),  35 -- 49 (2014)

\bibitem{mne}
Gramfort, A., Luessi, M., Larson, E., Engemann, D., Strohmeier, D., Brodbeck,
  C., Parkkonen, L., H\"am\"al\"ainen, M.: {MNE} software for processing {MEG}
  and {EEG} data. NeuroImage  86(0),  446 -- 460 (2014)

\bibitem{gramfort-etal:2013}
Gramfort, A., Strohmeier, D., Haueisen, J., H\"am\"al\"ainen, M., Kowalski, M.:
  Time-frequency mixed-norm estimates: Sparse {M/EEG} imaging with
  non-stationary source activations. NeuroImage  70(0),  410 -- 422 (2013)

\bibitem{guittet2002extended}
Guittet, K.: Extended kantorovich norms: a tool for optimization. Tech. Rep.
  4402, INRIA (2002)

\bibitem{hanin1999extension}
Hanin, L.: An extension of the kantorovich norm. Contemporary Mathematics  226,
   113--130 (1999)

\bibitem{henson-etal:11}
Henson, R.N., Wakeman, D.G., Litvak, V., Friston, K.J.: A parametric empirical
  bayesian framework for the {EEG/MEG} inverse problem: generative models for
  multisubject and multimodal integration. Front. in Human Neuro.  5(76) (2011)

\bibitem{Joshi2004S151}
Joshi, S., Davis, B., Jomier, M., Gerig, G.: Unbiased diffeomorphic atlas
  construction for computational anatomy. NeuroImage  23(0),  151--160 (2004)

\bibitem{kantorovich1958space}
Kantorovich, L., Rubinshtein, G.: On a space of totally additive functions,
  vestn. Vestn Lening. Univ.  13,  52--59 (1958)

\bibitem{kanwisher-etal:97}
Kanwisher, N., Mcdermott, J., Chun, M.M.: The fusiform face area: A module in
  human extrastriate cortex specialized for face perception. Journal of
  Neuroscience  17,  4302--4311 (1997)

\bibitem{pele2008linear}
Pele, O., Werman, M.: A linear time histogram metric for improved sift
  matching. In: European Conference on Computer Vision, ECCV 2008, pp. 495--508
  (2008)

\bibitem{pinel2007}
Pinel, P., Thirion, B., Meriaux, S., Jobert, A., Serres, J., Le~Bihan, D.,
  Poline, J., Dehaene, S.: Fast reproducible identification and large-scale
  databasing of individual functional cognitive networks. BMC neuroscience  8,
  ~91 (2007)

\bibitem{rabin2012}
Rabin, J., Peyr\'e, G., Delon, J., Bernot, M.: {W}asserstein barycenter and its
  application to texture mixing. In: Scale Space and Variational Methods in
  Computer Vision, Lecture Notes in Computer Science, vol. 6667, pp. 435--446.
  Springer (2012)

\bibitem{rubner1997earth}
Rubner, Y., Guibas, L., Tomasi, C.: The earth mover’s distance,
  multi-dimensional scaling, and color-based image retrieval. In: Proceedings
  of the ARPA Image Understanding Workshop. pp. 661--668 (1997)

\bibitem{scherg-etal:85}
Scherg, M., Von~Cramon, D.: {Two bilateral sources of the late AEP as
  identified by a spatio-temporal dipole model.} Electroencephalogr Clin
  Neurophysiol  62(1),  32--44 (Jan 1985)

\bibitem{thirion-etal:2007}
Thirion, B., Pinel, P., M\'eriaux, S., Roche, A., Dehaene, S., Poline, J.B.:
  {Analysis of a large fMRI cohort: Statistical and methodological issues for
  group analyses}. NeuroImage  35(1),  105 -- 120 (2007)

\bibitem{villani09}
Villani, C.: Optimal transport: old and new, vol. 338. Springer Verlag (2009)

\bibitem{wipf-etal:07}
Wipf, D., Ramirez, R., Palmer, J., Makeig, S., Rao, B.: Analysis of empirical
  bayesian methods for neuroelectromagnetic source localization. In: Proc.
  Neural Information Processing Systems (NIPS) (2007)

\end{thebibliography}

\end{document}